\def\year{2020}\relax
\documentclass[letterpaper]{article} 
\usepackage{aaai20}  
\usepackage{times}  
\usepackage{helvet} 
\usepackage{courier}  
\usepackage[hyphens]{url}  
\usepackage{graphicx} 
\usepackage{graphicx}  
\usepackage{dirtytalk}
\title{\say{It’s Unwieldy and It Takes a Lot of Time.}\\ Challenges and Opportunities for Creating Agents in Commercial Games}
\author{Mikhail Jacob,  
Sam Devlin,  
Katja Hofmann\\ 
Microsoft Research (Cambridge, UK)\\
\{t-mijaco, sam.devlin, katja.hofmann\}@microsoft.com 
}
\begin{document}

\maketitle

\begin{abstract}
Game agents such as opponents, non-player characters, and teammates are central to player experiences in many modern games. As the landscape of AI techniques used in the games industry evolves to adopt machine learning (ML) more widely, it is vital that the research community learn from the best practices cultivated within the industry over decades creating agents. However, although commercial game agent creation pipelines are more mature than those based on ML, opportunities for improvement still abound. As a foundation for shared progress identifying research opportunities between researchers and practitioners, we interviewed seventeen game agent creators from AAA studios, indie studios, and industrial research labs about the challenges they experienced with their professional workflows. Our study revealed several open challenges ranging from design to implementation and evaluation. We compare with literature from the research community that address the challenges identified and conclude by highlighting promising directions for future research supporting agent creation in the games industry.
\end{abstract}

\section{Introduction}

\begin{figure}[t]
\centering
\includegraphics[width=1.0\linewidth]{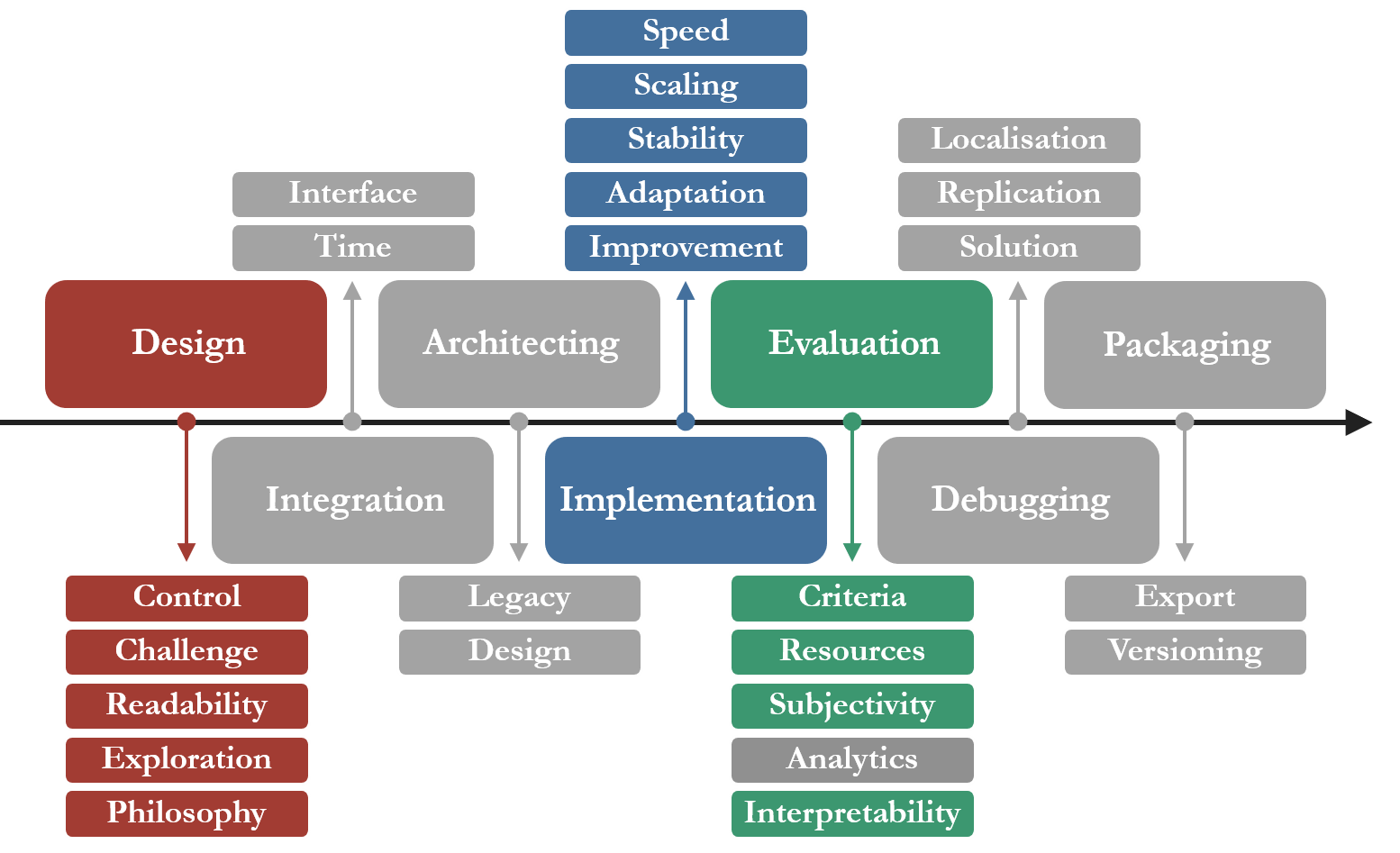}
\caption{A summary of themes surfaced by our interviews with game industry professionals about their workflows while creating game agents. This paper covers insights from the Design, Implementation, and Evaluation themes.} 
\label{themes}
\end{figure}

Game agents are a vital part of many games where they form a core part of the player experience, with the potential to elevate or break it. 
For example, the Xenomorph enemy NPC in Alien: Isolation \cite{creative_assembly_2014} is a singular source of tension and terror that was central to its success \cite{houghton_2014}. In contrast, Aliens: Colonial Marines \cite{gearbox_software_2013} received negative reviews due to the oblivious, disinterested enemies \cite{sterling_2013}.

These game agents are most commonly created in the games industry using techniques such as finite state machines, behaviour trees, utility systems, or planners \cite{rabin2015game}. More recently, game agents created with deep reinforcement learning (RL) and imitation learning (IL) have been demonstrated for commercial games as third-party bots for DotA 2 \cite{berner2019dota} and StarCraft II \cite{vinyals2019grandmaster}, as well as shipped creature NPCs for Source of Madness \cite{fornander_nilsson_2020}. As the landscape of AI approaches used in the game industry evolves, researchers have the opportunity to drive the inclusion of powerful RL and IL techniques through high-impact applied research by understanding what challenges and opportunities exist in the industry. The lessons learned and best practices cultivated within the industry over decades creating agents are also a valuable resource for researchers.

This paper synchronises expectations and illuminates opportunities as a foundation for shared progress between researchers and practitioners. To do so, we conducted an in-depth, qualitative research study systematically investigating the challenges that seventeen game agent creators from the games industry across AAA studios, indie studios, and industrial research labs actually experienced over the course of development. Using semi-structured interviews and thematic analysis, we extract key themes that especially affect the adoption of RL agent creation in commercial games. At the same time, these themes represent open opportunities for addressing issues with creating game agents by any method.

\section{Background}


We use the term `game agent' to refer to both game-playing `bots' that might replace human players as opponents or teammates, as well as non-player characters (NPCs) like enemies or companions that might serve more specific game roles. Game agent creators in the industry commonly have roles like `AI Engineer' or `AI Designer' that emphasise their focus on building features with code in contrast to designing the player experience. However, these distinctions are often blurred with engineers performing design tasks, technical designers taking on coding responsibilities, and some creators solely performing all AI-related tasks.

Game agent creators follow widely varying processes across organisations, agent techniques, team sizes, and individual preferences. However, they can be grouped at a higher level into the equivalent phases -- agent design and task specification, game integration, architecture design, agent implementation, evaluation and testing, debugging agent behaviour, as well as packaging and deploying agents (see Figure \ref{themes}). These phases are non-linear, iterative, and not always followed in the same order but do provide clear structure for comparing processes across these different groups. This high-level process is used throughout this article to organise insights from our interviews with game agent creators.

Game industry professionals and game AI researchers have published extensively about technological advances relevant to the industry \cite{rabin2015game,preuss2020games} and commercial game AI postmortems \cite{GDC_informa_plc_1988_2020}. These provide insight into the development of agents for individual commercial games or studios, but do not study challenges and opportunities at the meta-level that a cross-industry study such as this article can provide. Cross-industry resources like \cite{o2014developer} report findings from an ethnographic study of game companies but do not focus on game agents or game agent creation. 

\section{Methodology}

The purpose of our research is to identify open opportunities for supporting game agent creation. We focus here on the subjective experiences that a sample of game agent creators self-identified as important. Our methodology consisted of qualitative data collection using semi-structured interviews followed by thematic analysis. Real-world challenges that were identified in this manner were then compared with previously published literature. The results of this analysis were used to compile a list of open research challenges that could significantly impact game agent creation in industry.

\begin{table}
\resizebox{1.0\columnwidth}{!}
{
    \begin{tabular}{lllll}
    \textbf{ID} & \textbf{Org.} & \textbf{Role} & \textbf{Focus} & \textbf{Genre(s)} \\
    AAAAIEP01 & AAA & AI Engineer & Enemies & Action \\
    AAAAIEP02 & AAA & AI Engineer & MP Bots & Sports \\
    AAAAIEP03 & AAA & AI Engineer & MP Bots & RTS \\
    AAAAIDP04 & AAA & AI Designer & Enemies & FPS \\
    AAAAIEP05 & AAA & AI Engineer & MP Bots & FPS \\
    AAAAIEP06 & AAA & AI Engineer & Creatures & MMO \\
    INDAIEP07 & Indie & AI Engineer & Narrative & RTS, Sim \\
    LABRLEP08 & Lab & RL Engineer & Animation & Action \\
    INDAIEP09 & Indie & AI Engineer & MP Bots & Action \\
    AAAAIEP10 & AAA & AI Engineer & Enemies & FPS \\
    LABRLDP11 & Lab & RL Designer & Creatures & MMO \\
    LABRLEP12 & Lab & RL Engineer & MP Bots & Action \\
    LABRLEP13 & Lab & RL Engineer & MP Bots & Action \\
    LABRLEP14 & Lab & RL Engineer & MP Bots & FPS \\
    LABRLEP15 & Lab & RL Engineer & Tooling & Sim \\
    INDAIEP16 & Indie & AI Engineer & Narrative & RTS \\
    INDAIEP17 & Indie & AI Engineer & MP Bots & RTS
    \end{tabular}
}
\caption{Summary of Participants.
Abbreviations: Org. (Organisation), MP (Multiplayer), RTS (Real Time Strategy, FPS (First-Person Shooter), MMO (Massively Multiplayer Online), Sim (Simulation)}
\label{participants}
\end{table}


Data for this research was collected from $17$ professionals in the games industry, recruited using snowball sampling. Subjects were either part of a AAA studio ($7$ subjects), indie studio ($4$ subjects), or industrial research lab ($6$ subjects). 
Table \ref{participants} summarises participant details. Participants from game studios commonly used techniques such as finite state machines, behaviour trees, utility systems, and planners whilst industrial researchers used machine learning techniques such as RL and IL. Note that all researchers worked with commercial games but were expected to create proof of concept demonstrations rather than shipped products. We studied a small, rich sample of game creators, genres, and AI techniques, thus our study could be expanded in the future to a broader sample (e.g. including creative directors or business leadership).

Each data collection session consisted of a semi-structured interview focused on \emph{mapping out workflows}, \emph{challenges experienced}, \emph{tools used and desired}, as well as \emph{concerns and opportunities for machine learning} in one or more previous projects that involved creating game agents. For this article we focus on the portion of our data set corresponding to \emph{challenges experienced}. We used a qualitative methodology using semi-structured interviews with a small sample size instead of a large-scale survey in order to elicit open-ended and richly detailed responses that could be feasibly analysed by our research team. 
Interview data totalled over 21 hours of audio footage that were then transcribed into a corpus of textual transcripts and notes.


We analysed our data using thematic analysis following \citeauthor{braun2006using} \shortcite{braun2006using}. Using their recommended methodology, we iteratively `coded' (or annotated) words and phrases in the corpus describing instances and types of challenges experienced by the participants. We followed this with theme development and review, where codes were combined hierarchically into meaningful `themes' relating codes together and then reviewed for coherence. This resulted in categories of challenges that could be related to each other through a high-level process and compared across groups. This iterative process continued until thematic saturation.

\section{Results}

Thematic analysis of the interview data highlighted the many challenges experienced while creating game agents. We organise our results according to themes corresponding to high-level game agent creation processes (Figure \ref{themes}). Due to the page limit, we have selected the top three themes (Design, Implementation, and Evaluation) that arose from our analysis for detailed description below. In each thematic section, we present challenges described by participants followed by a discussion relating each challenge to published research or open opportunities for further contributions.

\subsection{Design}

Game agent designers focus on achieving a desired player experience through interactions with agents of different kinds. E.g., bots that are intended to replace human players are designed to take actions in-game mirroring human players, while NPCs are tightly designed to play a given role within the narrative or evoke a specific player affect. Thus the challenges designers described either related to workflow issues affecting their ability to impact player affect (like poor authorial control) or to aspects of player experience that were difficult to design for given their current capabilities (like behavioural readability, challenge matching, open-ended learning, and design philosophy changes).


\textbf{Control.} Seven participants echoed the importance of designer or authorial control over game agents. Participant AAAAIDP04 was frustrated with their lack of expressive designer control over character behaviours and reliance on engineering support for their NPC behaviour trees. Policy was to implement significant behavioural logic as opaque behaviour block code rather than logic in the tree. Designers had to request an engineer to write behaviour blocks for them or expose various block parameters and flags to designers in order to expressively differentiate characters. One of the few changes they could make themselves was to reorder behaviour blocks.
LABRLDP11 also described how RL affected designer control, saying, \say{it puts the kind of creative control interface between the designer and the game.} 


\textbf{Challenge:} 
Some participants found it challenging to design agents that provided an appropriate level of challenge for a given player's ability while remaining fun to play against. AAAAIEP03 noted that AI developers showed \say{a lack of planning for [creating a range of] AI difficulty levels.} Both AAAAIEP03 and AAAAIEP05 suggested that it was actually more difficult for them to design an effective bot for a novice player than for an experienced player. Designing different difficulty levels also required different strategies for bots compared to NPCs. For example, easy enemy NPCs in a shooter campaign might shoot slow projectiles enabling easier evasion, while this technique would break immersion in a deathmatch with bots adding additional noise to the bot targeting systems instead. INDAIEP17 suggested that the challenge was to create enemies that would do sub-optimal actions that were easy enough for the player to play against but were still interesting. 

\textbf{Readability:}
Designers encountered difficulties trying to successfully communicate to the player about agent's internal state subtly, without breaking immersion, in order to make its behaviour \say{readable}. AAAAIDP04 indicated that these issues often presented themselves because an enemy character's role in the game was to die in a way that didn't break player immersion and this required the player to be able to anticipate an enemy's moves or recognise when they had been spotted to take evasive action. AAAAIEP03 indicated that unreadable behaviour was considered unintelligent, saying, \say{if somebody else takes a look at a video of the AI and goes, \say{That just looks really stupid, \ldots why would your AI do that?}, as soon as they start asking that \say{why} question you know, okay, I \ldots have to explain why [the] AI is doing something[.]} Designers often used `barking' to verbally communicate about an agent's actions in order to increase readability. Barking was also used to fake complex reasoning for simple decision-making systems. 

\textbf{Exploration:}
LABRLDP11 described their design process using RL as iteratively setting a behavioural goal for the agent, running a training session, evaluating agent behaviour, and tweaking various parameters for the next training session. They felt that the process was overly task-focused and could use more open-ended exploration of behavioural possibilities, especially near the beginning of a project. They also found it challenging to design RL agents for sandbox games since the agent was being trained to solve a single task. The designer found it difficult to explore a range of experiences in one such game because they had to design a specific task, train the agent, test it in-game, and iterate each time, and the game was too dynamic and open-ended to make that process practical or replicable. More exploration would thus be welcome in both the RL training process and the kinds of experiences it can be applied to.

\textbf{Philosophy:}
Designer control is constrained by stronger agent autonomy provided by various AI techniques. Successfully designing with them requires a shift in agent design philosophy. AAAAIEP06 described how their designer colleague found it difficult to use their utility systems until they realised they were trying to force it to work like a more scripted system. AAAAIDP04 also explained the subtle shift in design philosophy needed to successfully design characters for systemic games rather than linear gameplay. For the latter, they would author specific sequences of actions for each level, while for systemic games they would give characters a set of capabilities and then design the levels and game systems to interact with those capabilities to elicit specific player experiences. 

Best practices for designing player experiences with RL agents in commercial games don't formally exist yet. However, perhaps successfully designing for reinforcement learning agents will involve similar design philosophy shifts to that required for designing agents for systemic and open-world games due to a focus on creating conditions for desirable behaviour to emerge in a trained agent rather than specifying those behaviours directly. As LABRLDP11 summarised about an agent designer's role with RL, \say{[It's] less about the details of the agents [and] it’s more about providing the props for today’s episode or the kind of dynamics of the environment [that produce interesting interactions].}

\textbf{Discussion:}
Enabling designers to blend authorial control with autonomy when creating agents to make the best of both paradigms remains an open challenge. Designers need to meaningfully control their exploration of the design space of possible game agents through quick design iteration. Imitation learning \cite{hussein2017imitation} is one way to intuitively exert designer control over what an agent learns by demonstrating desirable behaviours for the agent. Hybrid systems \cite{lee2018modular,neufeld2018hybrid} or interactive machine learning approaches \cite{frazier2019improving} could also be re-purposed for this goal. Rapid iteration would also be aided by agents that could quickly and meaningfully be adapted after training instead of retraining from scratch. Transfer learning \cite{taylor2009transfer} could be one potential direction towards addressing this opportunity. 

The balance between challenge and skill-level in a game can lead to apathy, boredom, anxiety, or optimal experience \cite{csikszentmihalyi1992optimal}. Dynamic difficulty adjustment \cite{hunicke2005case} is an established practice in the community for addressing this with its effect on player experience studied \cite{demediuk2019challenging}. Modern skill estimation and matchmaking approaches like TrueSkill \cite{herbrich2007trueskill} and AlphaRank \cite{omidshafiei2019alpha} can add to its effectiveness.

Agent behaviour is difficult to design so that it can be easily `read' by players. 
Behavioural readability challenges await RL creators with added significance due to the additional autonomy their agents possess. The use of barking in commercial game AI to address this problem could potentially be used for RL agents as well. \citeauthor{ehsan2019automated} \shortcite{ehsan2019automated} describe a method to generate natural language rationales for agent behaviour. Their approach, similarly to barking, generates rationales that are useful for players though not necessarily accurate or causal. More research into communication about agent decision-making would allow designers to show apparent intelligence for their agents. 

RL research often assumes there is a single specific task. This makes it challenging to train agents for open-ended games. Open-ended learning \cite{DoncieuxOpenEndedLearning2018}, continual learning \cite{ParisiContinualLifelong2018}, and multi-task learning \cite{ZhangASurvey2017} all aim at generalising agent training beyond a single task. 
Intrinsic motivation \cite{AubretIntrinsicMotivation2019} has also been used to aid in enabling agents to act autonomously without explicit goals, both in the RL and wider game AI communities. 
The continued growth of these research topics could enable RL to be effectively used in the industry for open-ended games.


\subsection{Implementation}

Most participants using both traditional game AI and RL encountered problems implementing and modifying game agents. These included insufficient training speed, scale, and stability for RL users early on and problems with agent adaptation and improvement for commercial game AI users over time. These differences highlight both the inherent technological challenges as well as the relative maturity in workflows and infrastructure between the two approaches.

\textbf{Speed:}
All the RL engineers interviewed noted the large number of environmental samples needed to see even early results from training agents in complex commercial video games. In contrast to the simple games frequently used in RL research, the AAA video games that the RL engineers were working with could not be sped up due to compute limitations and game physics degrading when sped up. This resulted in large delays evaluating training results in order to iterate on agent architecture or training parameters. LABRLEP14 discussed how an initial obstacle was the low sample throughput, \say{You run your first experiment, and then you realise that your agent is not going to learn anything because it can collect samples at five times a second, and to learn anything meaningful you need a couple of millions of samples [sic] which means that you need to run your experiment for a couple of weeks for it to learn anything.} LABRLDP11 described each experimental iteration taking days rather than hours because their training infrastructure at the time only allowed them to proceed serially and thus restricted the number of design ideas they could prototype.

\textbf{Scaling + Stability:}
The RL engineers approached the problem of long training times by scaling up compute infrastructure for training, using distributed training. In addition to scaling up, the engineers had to mitigate significant stability issues within the game builds and during training. The in-development game builds provided to the engineers had bugs that often blocked training or reduced the set of capabilities that an agent could be trained to use. This included memory leaks, network synchronisation bugs, and player attacks not working. Training infrastructure development also presented many challenges. Kubernetes was used initially to scale up training but proved too complex for reliable iteration in the long run. A distributed RL training framework (RLLib \cite{liang2018rllib}) was adopted as a replacement. While a significant improvement over the Kubernetes-based workflow, this also produced training instabilities due to underlying bugs. Finally, there were also instabilities caused by VM provisioning and failures in the cloud services provider that needed to be worked around.

\textbf{Adaptation:}
Many commercial game AI creators who participated in the study described the challenges encountered while adapting an existing agent to a new scenario. Participants highlighted the inappropriate behaviour displayed when characters were used in scenes for which they weren't designed. They would then need special-case scripting to fix the emergent bugs. AAAAIEP01 remarked how a large, boss monster was made playable on a map designed for small creatures and that led to boring fights where the monster looked stupid because it would just stand there unable to reach and attack the player. Custom scripting had to be added on top of the agent's behaviour tree to make it move away from a player's hiding spot to lure the player out before reverting to its normal behaviours. A similar experience adapting characters was echoed by AAAAIEP10 within a different shooter and was compounded by their inability to modify behaviour trees from existing characters for new scenarios even with significant overlaps. They also worked on implementing planners and observed that there were problems replicating successful behaviours when moving from small-scale gray box levels to fully developed levels due to differences in physical level of detail between the two.

\textbf{Improvement:}
Commercial game agent creators also agreed broadly that their current approaches to improving performance with AI techniques that required tuning many numeric parameters or parameter curves were extremely tedious and opaque to tune at scale. AAAAIEP05 characterised the opaque process of tuning their bot utility system as: \say{You really need to know what all of those [parameters] do, because \ldots if you do one thing with one set of inputs and you don’t do a corresponding action with other inputs, you’ll get into dangerous behaviour accidentally very quickly. There’s a very high barrier to entry[.]} AAAAIEP06 also found it challenging to scale up utility systems from tens of different creatures to hundreds of different creatures required for their MMO. It was a challenge to ensure consistent prioritisation of behaviours by the utility system for each agent. INDAIEP09 added that understanding how to tune numeric parameters in code to improve something qualitative like agent behaviour was a serious challenge for them.

\textbf{Discussion:}
Training speed, stability, and sample efficiency are known challenges experienced by deep RL practitioners. However, due to orders of magnitude difference in scale between RL research task environments and commercial games, it is particularly important to address these crucial issues during planning. \citeauthor{barriga2019improving} \shortcite{barriga2019improving} describes similar challenges while integrating deep RL into a commercial RTS game along with the compromises required to adopt it successfully. \citeauthor{jia2020mastering} \shortcite{jia2020mastering} recently described the challenges and lessons learned working to integrate RL with a commercial basketball games. \citeauthor{shih_teng_nilsson_2020} \shortcite{shih_teng_nilsson_2020} also described similar insights as an indie studio using RL to control enemy creature NPCs in their upcoming game. RL research has thus shown an emerging trend to focus directly on the practical issues emerging from integrating deep RL with commercial games. These practical insights need to be cemented into best practices ensuring strongly reproducible results \cite{Pineau2020ImprovingRI} evaluated by robust metrics \cite{rl_reliability_metrics} before they can be confidently invested in and adopted by game studios.
    
Poor generalisation in RL agents across levels or in interactions with other agents is another well-known challenge. There is an emerging RL research trend focusing on generalisation \cite{Cobbe2019QuantifyingGI}. However, this topic has a long history of study within the game AI research community through the GVGAI platform \cite{perez2016general} and domain randomisation using procedural content generation (PCG) \cite{risi2020increasing}. 
Observations about the lack of agent adaptation presented earlier, highlight the challenges that RL agent creators are likely to face as the technology is embraced by commercial games at scale.
    
Improving agent performance by tuning numeric hyperparameters is a challenge for RL practitioners similarly to the commercial game AI creators who noted the challenge. Automated searches for hyperparameter configurations are commonly used in RL (and more generally ML). These include default grid searches, as well as more recent approaches such as meta-learning \cite{xu2018meta}. Automated parameter tuning techniques would be generally useful for commercial game agent creators using any agent architectures with many numeric parameters.

\subsection{Evaluation}

Participants working with both commercial game AI and RL broadly described evaluation and testing of game agent behaviour as a significant challenge for them. At a high level, this was due either to resources needed for testing or the subjectivity of evaluation. RL users also pointed to the lack of interpretability in their neural network models.

\textbf{Criteria:}
Agent performance was evaluated against several kinds of criteria during prototyping and production. This included showing skills at navigation, attacking, and evasion as measured by game-specific analytics like win rates or K-D (kill - death) ratios. This also included effects on player experience, such as whether it was a threatening presence at close range. Several developers aimed at human-like behaviour. LABRLEP14 pointed to apparent intelligence, saying, \say{a bot that would \ldots do things that make sense, \ldots not necessarily optimal but make sense, would be something useful[.]} INDAIEP09 evaluated agents by comparing agent actions to ideal human actions. One participant warned from experience, however, that imitating human behaviour without guardrails could result in toxic agent behaviours and affect long-term player feedback negatively despite initially temporarily boosting reviews. AAAAIDP04 emphasised player experience over all else for agent creation saying that the most important criterion was, \say{Is it fun?}

Several commercial game agent creators described compute performance as a practical metric due to strict compute budgets at runtime. AAAAIEP05 decided to use utility systems as opposed to neural networks for their agent because of this concern. AAAAIEP02 described their compute budget as three milliseconds to generate actions for twenty-four opponents on seven-year-old hardware. INDAIEP17 added that they had to spend precious time optimising agent performance as a small team working on consoles.

\textbf{Resources:}
Many commercial agent creators felt that testing agent behaviours ate into their design or development responsibilities. Lacking a formal QA team, Indie creators like INDAIEP07 and INDAIEP17 had no option but to test everything themselves. Even with large QA teams, AAAAIDP04 observed that designer playtesting had no substitute since there was a risk of QA confusing unusual but desirable behaviour with actual bugs. AAAAIEP03 suggested large-scale test automation as a powerful solution that competed with other production priorities to its detriment when a game was already on fire. However, some participants had adopted limited amounts of test automation for narrowly-focused tasks like smoke checking (INDAIEP07, AAAAIEP06), creature navigation evaluation (AAAAIEP06), and bot skill evaluation (AAAAIEP01).

Two RL participants did not have QA available to them and described requiring regular manual testing over the course of a training session. LABRLEP14 labelled the human effort needed to evaluate each iteration of an agent tedious. At interview time, LABRLEP12 had succeeded in partially automating the collection of evaluation statistics during training. Their RL agent had to be tested against scripted agents or human players to measure progress.

\textbf{Subjectivity:}
Evaluating an agent's behaviour was challenging due to subjectivity in behavioural metrics. INDAIEP17 noted the difficulty of evaluating behaviours during playtesting except for when the player's crew was not carrying orders correctly. Multiple AI engineers noted the uncertainty of determining whether changes made to the agent had made a tangible difference. For instance, AAAAIEP06 observed that tuning utility curves didn't immediately show a difference in agent behaviour. They also described how evaluating character behaviours was made more difficult by their game's open-endedness and emergence. Their previous strategy, behavioural unit tests like whether an agent had jumped at all since starting, was infeasible for their current game for the same reason.




\textbf{Interpretability:}
3 RL creators observed that predicting a trained agent's behaviour from graphs of training statistics was challenging. LABRLDP11 clarified that training graphs were useful for well-defined tasks with reward structures exactly matching a desired outcome. They explained that suspiciously high returns were obtained by `reward hacking,' i.e. an agent finding an exploit to earn rewards without completing the task. LABRLEP13 also noted that graphed metrics did not necessarily correspond to improved player experience. Instead, they preferred to ask, \say{Does it feel fun to play against? Or does it look like it’s doing the right thing?} Therefore, they would also regularly playtest runs of the games generating recordings of the agent's attempt for documenting and evaluating its behaviour. However, roll outs still required manual evaluation of recordings at scale. 

\textbf{Discussion:}
Both game AI \cite{isbister2006better}  and RL \cite{hussein2017imitation} research communities have studied the production and evaluation of human-like behaviour in agents. However, a game agent's effect on player experience is perhaps the most important measure of its success as described by both commercial game AI and RL creators. Therefore, as RL is used to create agents for commercial games, creators must ensure that the effect of these agents on player experience is evaluated. Game user research \cite{el2016game}, as it is conducted by industry and game AI researchers, must be prioritised using quantitative metrics, physiological sensors, interviews, or questionnaires.

Commercial adoption of RL also hinges on its ability to perform within the developer's compute budget. This includes showing empirical guarantees on response times for agent decision-making and memory usage across a large number of agents. \citeauthor{powley2017memory} \shortcite{powley2017memory} demonstrated MCTS in card and board games with a small memory footprint. Similarly, RL research needs to demonstrate empirical bounds on a range of compute performance metrics to build trust in the technology. Methods that are shown to work within tight computational constraints are more likely to be adopted by the commercial games industry.

Extensively testing and evaluating agent behaviours is a fundamental yet resource-intensive activity for game agent creators. Though agents are often used to test aspects of a game such as level design \cite{holmgard2018automated}, agent behavioural test automation is still limited in the game industry and for easily measured metrics. Similar to smaller indie creators, RL research labs typically don't have QA personnel dedicated to evaluating trained agents manually. Therefore, automated agent testing would be particularly beneficial to RL agent creators. 
Standardised testing suites, perhaps across comparable game genres, could enable automated testing more broadly for RL. However, agents will need to generalise successfully from test environments to game environments for this approach to show results.

Recent techniques from the explainable AI \cite{samek2019explainable} research community applied to RL could be used to address the challenge of interpreting and explaining decisions generated by the agent's neural network model. Saliency maps can be used to evaluate the relevance of specific observations to the final action generated by the model \cite{Atrey2020Exploratory} or a generative model could be trained to interrogate the model about critical states \cite{Rupprecht2020Finding}. Another possibility is to re-purpose techniques from game analytics research \cite{el2016game} for understanding human player experience like player progression metrics or play style clustering to evaluate agent behaviour.
Finally, limited human playtesting could be conducted against the most promising agents. This would provide convincing evidence about the value that agents provided to players in addition to confirming that agents were indeed fun to play with. 

\section{Conclusion}

\begin{table}
\centering
\resizebox{1.0\columnwidth}{!}
{
    \begin{tabular}{|p{\columnwidth}|}
    \hline
    \\
    \multicolumn{1}{|c|}{\textbf{Research Opportunities for Agents in Commercial Games}} \\[-1ex]
    \begin{itemize}
        \item Give more authorial control to agent designers.
        \item Player skill evaluation for better challenge matching
        \item Design agents for greater behavioural readability.
        \item Agent design needs to accommodate open-endedness.
        \item Reproducible, practical insights and best practices from integrating agents with commercial games are crucial.
        \item Agents created at scale need to generalise and adapt.
        \item Automated parameter tuning would speed up iteration.
        \item Prioritise player experience and compute performance in addition to human-likeness.
        \item Automated testing would allow focused creative iteration.
        \item Agent models need interpretability tools and techniques.
    \end{itemize} \\
    \hline
    \end{tabular}
}
\caption{A summary of highlighted research opportunities for game agent creation in commercial games}
\label{recommendations}
\end{table}

Machine learning techniques are an exciting material for creating game agent behaviour. While compelling demonstrations of RL and IL within commercial games have started to appear more frequently, mass adoption by commercial game creators will require significant technical innovation to build trust in these approaches. This article provides a shared foundation for progress between researchers and practitioners by highlighting challenges that impede the adoption of RL and IL in the games industry. Our results offer opportunities for researchers to directly shape the adoption of these technologies, while remaining informed by the decades of experience creating game agents in the industry.

Our interviews with game industry professionals showcased the challenges they experienced throughout their current game agent creation workflows whether using traditional game AI approaches or RL. We described challenges thematically organised into the \emph{design}, \emph{implementation}, and \emph{evaluation} phases of a high-level game agent creation pipeline. We discussed opportunities in each theme (summarised in Table \ref{recommendations}) for RL research and new approaches to creating game agents more broadly. We expect that our recommendations will foster research to lower the technical barrier through replicable best practices, mature the agent creation pipeline with rapid iteration and exploration at scale, as well as encourage researchers to make impacting player experience a first-class goal.


{\small
\bibliography{aiide20-paper-references}
\bibliographystyle{aaai}}
\end{document}